\documentclass{bmvc2k}

\usepackage{amsmath}
\usepackage{amssymb}
\usepackage{booktabs}
\usepackage{multirow}
\usepackage{bbding}

\usepackage[normalem]{ulem}
\useunder{\uline}{\ul}{}

\title{Prompting Visual-Language Models for Dynamic Facial Expression Recognition}

\addauthor{Zengqun Zhao}{zengqun.zhao@qmul.ac.uk}{1}
\addauthor{Ioannis Patras}{i.patras@qmul.ac.uk}{1}

\addinstitution{School of Electronic Engineering and Computer Science,\\
Queen Mary University of London,\\
London, E1 4NS, UK
}

\runninghead{ZHAO and Patras}{Prompting Visual-Language Models for DFER}

\begin{document}
\maketitle
\begin{abstract}
This paper presents a novel visual-language model called DFER-CLIP, which is based on the CLIP model and designed for in-the-wild Dynamic Facial Expression Recognition (DFER). Specifically, the proposed DFER-CLIP consists of a visual part and a textual part. For the visual part, based on the CLIP image encoder, a temporal model consisting of several Transformer encoders is introduced for extracting temporal facial expression features, and the final feature embedding is obtained as a learnable "class" token. For the textual part, we use as inputs textual descriptions of the facial behaviour that is related to the classes (facial expressions) that we are interested in recognising -- those descriptions are generated using large language models, like ChatGPT. This, in contrast to works that use only the class names and more accurately captures the relationship between them. Alongside the textual description, we introduce a learnable token which helps the model learn relevant context information for each expression during training. Extensive experiments demonstrate the effectiveness of the proposed method and show that our DFER-CLIP also achieves state-of-the-art results compared with the current supervised DFER methods on the DFEW, FERV39k, and MAFW benchmarks. Code is publicly available at \url{https://github.com/zengqunzhao/DFER-CLIP}. 

\end{abstract}
\section{Introduction}
Facial expression is an important aspect of daily conversation and communication \cite{darwin1998expression}. Because of its application in various fields, such as human-computer interaction (HCI) \cite{duric2002integrating}, driving assistance \cite{wilhelm2019towards}, and mental health \cite{bishay2019schinet,foteinopoulou2021estimating}, facial expression recognition (FER) attracts increasing attention from researchers in psychology, computer science, linguistics, neuroscience, and related disciplines \cite{corneanu2016survey}. When using a discrete emotion model \cite{ekman1971universals}, FER aims to classify an image or video sequence into one of several basic emotions, i.e., neutral, happiness, sadness, surprise, fear, disgust, and anger. Traditional facial expression recognition approaches have mainly focused on static images or video frames, which do not capture the temporal dynamics of facial expressions \cite{li2020deep,zhao2021learning}. However, the recognition of dynamic facial expressions involves capturing the changes in facial movements over time, which can provide more information for accurate emotion recognition. Therefore, the recognition of dynamic facial expressions has become an increasingly important research area within the field of computer vision and affective computing \cite{li2020deep,wang2022systematic}.

Early studies for dynamic facial expression recognition (DFER) are mainly focused on lab-controlled conditions \cite{li2020deep}, in which the faces are all frontal and have no occlusions. However, in the real world, facial expressions always occur accompanied by variations such as illumination, non-frontal pose, and occlusion. Therefore, works on in-the-wild DFER are getting popular in recent years \cite{jiang2020dfew,zhao2021former,wang2022ferv39k,liu2022mafw,liu2023expression}. The main focus is trying to learn discriminative and robust spatial-temporal feature representations for the DFER task. These methods can be roughly categorized into 3DCNN-based \cite{jiang2020dfew,lee2019context}, CNN-RNN-based \cite{baddar2019mode,liu2021video,wang2022dpcnet}, and Transformer-based \cite{zhao2021former,foteinopoulou2022learning,li2023intensity,liu2023expression}, in which Transformer-based methods have achieved state-of-the-art performance. In addition, many in-the-wild datasets were proposed in recent years for DFER \cite{jiang2020dfew,liu2022mafw,wang2022ferv39k}. However, the high cost of labelling facial expression data limits the amount of training data available for DFER models. This can have an impact on the model's performance, especially if the training data does not fully capture the diversity of facial expressions that may occur in the real world.

\begin{figure*}[!t]
	\centering
	\includegraphics[scale=0.42]{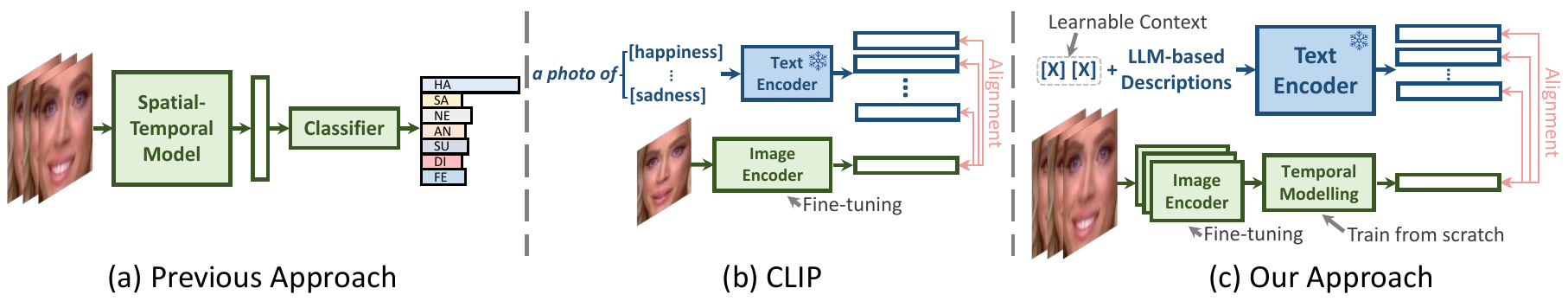}
	\caption{Illustration of the (a) previous approach for DFER, which relies on a classifier to predict the facial expression label. (b) Standard vision-language model CLIP. (c) Based on the CLIP, we propose a DFER-CLIP, which further models temporal facial features and incorporates fine-grained descriptors.}
	\label{f1}
\end{figure*}

Recently, vision-language pre-training (V-LP) models such as CLIP \cite{radford2021learning} have emerged as promising alternatives for visual representation learning. The main idea of the V-LP models is to align large connections of images and raw texts using two separate encoders, so as to learn semantic information between visual and textual data. The success of V-LP models has led to a growing interest in using them for various downstream computer vision tasks, such as video understanding \cite{ju2022prompting,rasheed2022fine}, image synthesis \cite{patashnik2021styleclip}, and semantic segmentation \cite{ma2022open}. Given the powerful representation learning capabilities of V-LP models, a relevant question is how to best exploit their potential for the DFER task. One proposed approach is to fine-tune the image encoder of a V-LP model on DFER datasets. However, this approach faces two challenges that must be addressed. Firstly, while the original CLIP model can recognize objects and scenes in images, it may not be as effective in recognizing subtle facial expressions, which require more fine-grained descriptors and modelling of the similarities between expressions. Secondly, learning robust temporal facial features to understand emotions is crucial for DFER. Unfortunately, the standard CLIP visual encoder encodes static images and therefore lacks ways of capturing temporal information.

\begin{figure*}[!t]
	\centering
	\includegraphics[scale=0.67]{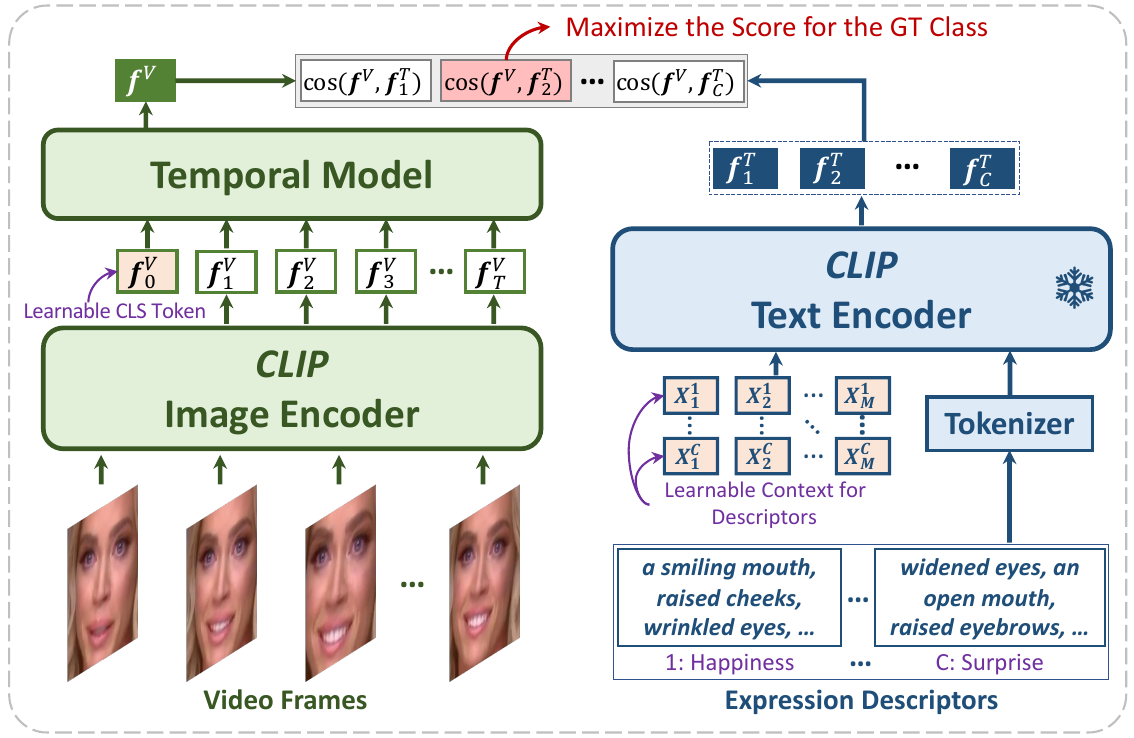}
	\caption{The structure of the proposed DFER-CLIP model. cos() denotes cosine similarity. M denotes the learnable context number. C denotes expression classes.}
	\label{f2}
\end{figure*}

To address these challenges, based on the CLIP model, we propose a novel architecture, namely the DFER-CLIP model. An overview of the differences between DFER-CLIP with previous approaches and CLIP is shown in Fig.~\ref{f1} and the detailed architecture is shown in Fig.~\ref{f2}. The proposed method mainly consists of a visual part and a textual part. Regarding the visual part, based on the CLIP image encoder, we introduce a temporal model, which consists of several Transformer encoders \cite{dosovitskiy2020image}, for modelling the temporal facial features. The final video-level facial features are obtained by a learnable class token. Regarding the textual part, considering that different facial expressions have both common properties and unique or special properties at the level of local behaviour \cite{zhong2012learning,zhao2021robust}, we utilize descriptions related to facial behaviour instead of class names for the text encoder. As a result, the text embedding can provide more detailed and precise information about the specific movements or positions of muscles involved in each expression. Furthermore, inspired by CoOp \cite{zhou2022learning}, we adopt the learnable prompt as a context for descriptors of each class -- this does not require experts to design context words and allows the model to learn relevant context information for each expression during training. To evaluate our DFER-CLIP model, we conducted experiments on three datasets. The results show that the temporal model can clearly enhance performance, and adopting the fine-grained expression descriptions with learnable context is superior to class-level prompts. Furthermore, compared with the current supervised DFER methods, the proposed DFER-CLIP achieves state-of-the-art results on DFEW, FERV39k, and MAFW benchmarks.

\section{Related Work}
\subsection{DFER in the Wild}
For in-the-wild DFER, deep learning-based methods have achieved superior performance to methods using hand-crafted features in recent years \cite{li2020deep}. Those methods are mainly focusing on learning robust spatial-temporal facial expression features, and can be roughly categorized into 3DCNN-based \cite{jiang2020dfew,lee2019context}, CNN-RNN-based \cite{baddar2019mode,liu2021video,wang2022dpcnet}, and Transformer-based \cite{zhao2021former,foteinopoulou2022learning,li2023intensity,liu2023expression} methods. To incorporate more modalities for emotion understanding, multi-modal learning is also prevalent in DFER in the wild \cite{chen2016facial,mittal2020m3er}. Zhao \textit{et al.} \cite{zhao2021former} first introduced the Transformer into DFER and achieved state-of-the-art results. After that, many transformer-based methods \cite{ma2023logo,li2022nr,li2023intensity} were proposed to further enhance the performance. In the latest work, Li \textit{et al.} \cite{li2023cliper} proposed a model called CLIPER that uses visual language models for the FER task, achieving state-of-the-art performance. However, the model is limited in its ability to capture temporal information for dynamic FER and uses a two-stage training paradigm instead of end-to-end training. By contrast, our DFER-CLIP model is capable of learning spatial-temporal features and is trained in an end-to-end manner. 

\subsection{Visual-Language Models}
Vision-language models have made significant progress in zero-shot performance through contrastive learning on pairs of images and text \cite{radford2021learning,zhou2022learning,zhou2022conditional,menon2022visual,ju2022prompting,rasheed2022fine,oldfield2023parts}. This approach leverages a vast and virtually limitless supply of image-text pairs collected from the internet, making the process both highly efficient and cost-effective. The most recent vision-language learning frameworks CLIP \cite{radford2021learning} and ALIGN \cite{jia2021scaling} benefit from 400 million and 1.8 billion image-text pairs collected from the Internet. These large-scale vision-language models are then applied to various downstream computer vision tasks and achieve stage-of-the-art performance \cite{patashnik2021styleclip,ni2022expanding,nag2022zero,ju2022prompting,rasheed2022fine}. For the visual-language modes, the researchers usually utilize a prompt (e.g., “a photo of a”) along with the name of the class to generate a text descriptor for each class. Zhou \textit{et al.} \cite{zhou2022learning} introduced a learnable continuous prompt method named context optimization (CoOp), which can adaptively adjust according to different datasets. Furthermore, Zhou \textit{et el.} \cite{zhou2022conditional} proposed conditional context optimization (CoCoOp) based on CoOp, which can generate a prompt conditioned on each image and make better inferences on unseen classes. Li \textit{et al.} \cite{li2022blip} proposed a new V-LP model named BLIP which achieves state-of-the-art performance on both understanding-based and generation-based tasks. BLIP trains a multimodal mixture of encoder-decoder models on large-scale noisy image-text pairs. In the latest work, Li \textit{et al.} \cite{li2023scaling} proposed a simple and more efficient method for training CLIP, named Fast Language-Image Pre-training (FLIP). They randomly mask out and remove a large portion of image patches during training, allowing more image-text pairs can be used for training under the same overheads. Our method is designed based on the CLIP model.

\section{Methods}
\subsection{Overview}
The structure of the DFER-CLIP is shown in Fig. \ref{f2}. As we can see, the proposed DFER-CLIP consists of a visual part and a textual part. Regarding the visual part, based on the CLIP image encoder, we introduce a temporal model, which consists of several Transformer Encoders, for modelling the temporal facial features. The final video-level facial features are obtained by a learnable class token. Regarding the textual part, we utilize descriptions related to facial behaviour instead of class names for the text encoder. Furthermore, we adopt the learnable prompt as a context for descriptors of each class, which does not require experts to design context words and allows the model to learn relevant context information for each expression during training.  We will first introduce the detail of the DFER-CLIP model and then show how we build textural descriptions for facial expressions.

\subsection{DFER-CLIP}
Regarding the visual part, on top of the CLIP image encoder, the temporal model consisting of several ViT encoder layers is adopted for modelling the temporal relationship. Each encoder also consists of a multi-headed self-attention and feed-forward network, which are all trained from scratch. The frame-level features are first learnt by the shared CLIP visual encoder. Then all of the frame-level features along with an additional learnable class token will feed into the temporal model, in which the learnable position embedding is added to encode the temporal position.

Concretely, given a facial video, we sample $ T $ frames of the size $ H \times W $ so as to form an input $ \textit{x} \in \mathbb{R}^{T \times 3 \times H \times W} $. For each frame $ \textit{x}_{i} $, we first utilize a shared CLIP image encoder $f(\cdot)$ to extract feature vectors $ \textbf{\textit{f}}^{v}_{i} \in \mathbb{R}^{L} $, where $ i \in \lbrace 1, 2, \cdots, T \rbrace $, $L$ is the length of the feature vectors. Then $T$ feature vectors will feed into a temporal model $g(\cdot)$ for learning temporal features, and the final visual representation $ \textbf{\textit{f}}^{V} \in \mathbb{R}^{L} $ can be obtained:
\begin{equation}
\begin{aligned}
\textbf{\textit{f}}^{v}_{i} = f(x_{i})
\end{aligned} 
\label{equ11}
\end{equation}
\begin{equation}
\begin{aligned}
\textbf{\textit{f}}^{V} = g(\textbf{\textit{f}}^{v}_{0}+\textbf{\textit{e}}_{0}, \textbf{\textit{f}}^{v}_{1}+\textbf{\textit{e}}_{1}, ..., \textbf{\textit{f}}^{v}_{T}+\textbf{\textit{e}}_{T})
\end{aligned} 
\label{equ12}
\end{equation}
where $\textbf{\textit{f}}^{v}_{0}$ is a special learnable vector for the class token, $\textbf{\textit{e}}$ represents a learnable position embedding added to encode the temporal position.

Regarding the textual part, we utilize descriptions related to facial behaviour instead of class names for the text encoder. Furthermore, we adopt the learnable prompt as a context for descriptors of each class. The form of the prompt is as follows:
\begin{equation}
\begin{aligned}
\textbf{\textit{P}}_{k} = [p]_{k}^{1} [p]_{k}^{2} \cdots [p]_{k}^{M} [Tokenizer(description)]_{k}
\end{aligned} 
\label{equ2}
\end{equation}
where $M$ is a hyperparameter specifying the number of context tokens, $ k \in \lbrace 1, 2, \cdots, C \rbrace $, $C$ is the number of the facial expression classes, and each $[p]_{k}^{m}$, $ m \in \lbrace 1, 2, \cdots, M \rbrace $, is a vector with the same dimension as word embeddings. Here, we adopt class-specific context where context vectors are independent of each description. By forwarding a prompt $\textbf{\textit{P}}_{k}$ to the text encoder $h(\cdot)$, we can obtain $C$ classification weight vector $ \textbf{\textit{f}}^{T}_{k} \in \mathbb{R}^{L}$ representing a visual concept:
\begin{equation}
\begin{aligned}
\textbf{\textit{f}}^{T}_{k} = h(\textbf{\textit{P}}_{k})
\end{aligned} 
\label{equ3}
\end{equation}

Then the prediction probability can be computed as:
\begin{equation}
\begin{aligned}
p(y = k | x) = \dfrac{exp(cos(\textbf{\textit{f}}^{V}, \textbf{\textit{f}}^{T}_{k})/\tau)}{\sum_{k^{\prime}=1}^{C} exp(cos(\textbf{\textit{f}}^{V}, \textbf{\textit{f}}^{T}_{k^{\prime}})/\tau)} 
\end{aligned} 
\label{equ4}
\end{equation}
where $\tau$ is a temperature parameter learned by CLIP and $cos(\cdot, \cdot)$ denotes cosine similarity.

During the training phase, the CLIP text encoder is fixed and we fine-tune the CLIP image encoder. The temporal model, learnable class token, and learnable context are all learned from scratch. DFER-CLIP is trained end to end and the cross-entropy loss is adopted for measuring the distance between the prediction and the ground-truth labels. 

\subsection{Building Descriptions}
Previous research has indicated that human facial expressions share some common properties while also possessing unique or distinctive features \cite{zhong2012learning}. This suggests that there are similarities in the movements of local facial muscles across different expressions. For example, the expressions of happiness and surprise both involve the raising of the eyebrows, and the expressions of sadness and anger both involve the lowering of the eyebrows and the furrowing of the forehead. Considering the fact that the CLIP text encoder learned semantic information from natural language text \cite{radford2021learning}, we propose taking the facial expression action description as the input for the text encoder. Instead of manually designing the facial expression description, we prompt a large language model such as ChatGPT to automatically generate descriptions based on contextual information. We prompt the language model with the input: \\
\indent \textit{Q: What are useful visual features for the facial expression of \{class name\}?}  \\
\indent \textit{A: Some useful visual features for facial expressions of \{class name\} include: ...} \\
All the generated descriptors of each facial expression class will be combined to form a comprehensive description. The descriptions for the classification of the seven basic emotions and of the classes in the datasets we used can be found in the \textit{supplementary material}.

\section{Experiments}

\subsection{Datasets}
 To evaluate our method, we conduct experiments on three popular DFER benchmarks.
 
 \textbf{DFEW} \cite{jiang2020dfew} dataset contains 11,697 in-the-wild video clips, and all the samples have been split into five same-size parts without overlap. Each video is individually annotated by ten annotators under professional guidance and assigned to one of the seven basic expressions (i.e., happiness, sadness, neutral, anger, surprise, disgust, and fear). The video clips are collected from over 1,500 movies worldwide, covering various challenging interferences, such as extreme illuminations, occlusions, and variant head pose. The 5-fold cross-validation is adopted as an evaluation protocol.
 
 \textbf{FERV39k} \cite{wang2022ferv39k} dataset contains 38,935 in-the-wild video clips, which is currently the largest in-the-wild DFER dataset. All the video clips were collected from 4 scenarios, which can be further divided into 22 fine-grained scenarios, such as crime, daily life, speech, and war. Each clip is annotated by 30 individual annotators and assigned to one of the seven basic expressions as DFEW. Video clips of all scenes are randomly shuffled and split into training (80\%), and testing (20\%) without overlapping.
 
 \textbf{MAFW} \cite{liu2022mafw} dataset contains 10,045 in-the-wild video clips, which is the first large-scale, multi-modal, multi-label affective database with 11 single-expression categories (anger, disgust, fear, happiness, sadness, surprise, contempt, anxiety, helplessness, disappointment, and neutral), 32 multiple-expression categories, and emotional descriptive texts. The 5-fold cross-validation is adopted as an evaluation protocol.

\subsection{Implementation Details}
\textbf{\textit{Training Setting}}: For the visual input, all the fixed 16 frames of sequence in our experiments follow the sampling strategy in \cite{jiang2020dfew,zhao2021former,ma2023logo,li2023intensity,li2022nr,li2023cliper}, and then resized to $224 \times 224$. The random resized crop, horizontal flipping, random rotation, and color jittering are used to avoid overfitting. Considering the high computational cost of video data, ViT-B/32-based CLIP \cite{radford2021learning} is employed in our DFER-CLIP. The maximum number of textual tokens is 77 (following the official CLIP design), and the temperature hyper-parameter $\tau$ is set to 0.01. Our models are trained on the Tesla A100 GPU based on the open-source PyTorch platform. The parameters are optimized via the SGD optimizer with a mini-batch size of 48. The initial learning rate for the CLIP image encoder, temporal model, and learnable prompt is set as $1\times10^{-5}$, $1\times10^{-2}$, and $1\times10^{-3}$, respectively. The MultiStepLR with milestones of \{30, 40\} and gamma of 0.1 is employed as a scheduler to adjust the learning rate. The DFER-CLIP model is trained for 50 epochs in an end-to-end manner. To obtain more stable and reliable results, we train models three times with different random seeds and then use the average as the final result.     

\textbf{\textit{Evaluation Metrics}}: Consistent with most previous methods \cite{jiang2020dfew,zhao2021former,ma2023logo,li2023intensity,li2022nr,li2023cliper}, the weighted average recall (WAR, i.e., accuracy) and the unweighted average recall (UAR, i.e., the accuracy per class divided by the number of classes without considering the number of instances per class) are utilized for evaluating the performance of methods.

\begin{table}[t]
\begin{center}
\small
\begin{tabular}{@{}c|c|cc|cccc@{}}
\toprule
\multirow{2}{*}{\begin{tabular}[c]{@{}c@{}}Number of the\\ Context Prompts\end{tabular}} &
  \multirow{2}{*}{\begin{tabular}[c]{@{}c@{}}Depth of the\\ Temporal Model\end{tabular}} &
  \multicolumn{2}{c|}{DFEW} &
  \multicolumn{2}{c|}{FERV39k} &
  \multicolumn{2}{c}{MAFW} \\ \cmidrule(l){3-8} 
   &              & UAR            & WAR            & UAR            & \multicolumn{1}{c|}{WAR}            & UAR            & WAR            \\ \midrule
4  & \XSolidBrush & 56.91          & \textbf{69.01} & 40.26          & \multicolumn{1}{c|}{\textbf{50.96}} & 38.03          & 50.62          \\
8  & \XSolidBrush & \textbf{57.39} & 69.00          & \textbf{40.64} & \multicolumn{1}{c|}{50.92}          & \textbf{38.51} & \textbf{50.91} \\
16 & \XSolidBrush & 57.32          & 68.96          & 40.22          & \multicolumn{1}{c|}{50.64}          & 37.98          & 50.40          \\ \midrule
8  & 1            & \textbf{59.61} & \textbf{71.25} & \textbf{41.27} & \multicolumn{1}{c|}{\textbf{51.65}} & \textbf{39.89} & \textbf{52.55} \\
8  & 2            & 58.87          & 70.92          & 40.41          & \multicolumn{1}{c|}{51.08}          & 39.13          & 52.10          \\
8  & 3            & 58.64          & 70.80          & 40.35          & \multicolumn{1}{c|}{50.98}          & 38.90          & 51.86          \\ \bottomrule
\end{tabular}
\end{center}
\caption{Evaluation of the learnable context prompt numbers \& the temporal model depths.}
\label{tab1}
\end{table}

\subsection{Ablation Analysis}
To evaluate the effectiveness of each component in our DFER-CLIP, we conduct exhaustive ablation analysis on all three benchmarks. Specifically, we will first demonstrate the effectiveness of the temporal model, the effect of the different number of context prompts, and the depth of the temporal model. We will then compare different training strategies, including classifier-based and text-based (classifier-free). Finally, we will compare different prompts for the DFER task.

\textbf{\textit{Evaluation of Temporal Model \& Context Prompts}}: As aforementioned, learning temporal facial features is crucial for video-based FER tasks. From Tab. \ref{tab1}, we can see that by adopting the temporal model, the UAR performance can be improved by 2.22\%, 0.63\%, and 1.38\%, and WAR performance can be improved by 2.25\%, 0.73\%, and 1.64\% on DFER, FERV39k, and MAFW datasets, respectively. In general, deeper models can get better performance, but in our DFER-CLIP, the best performance is obtained under the one-layer temporal model. This is because the temporal model is trained from scratch and may overfit if it is complex. This is also a consideration in the learnable context, in which more learnable vectors do not improve the results. We believe that increasing the learnable context number or temporal model depth will cause overfitting on the training data, resulting in a worse generalization performance on test data.  
\begin{table}[t]
\begin{center}
\footnotesize
\begin{tabular}{@{}cc|cc|cc|cc@{}}
\toprule
\multicolumn{2}{c|}{\multirow{2}{*}{Strategies}} &
  \multicolumn{2}{c|}{DFEW} &
  \multicolumn{2}{c|}{FERV39k} &
  \multicolumn{2}{c}{MAFW} \\ \cmidrule(l){3-8} 
\multicolumn{2}{c|}{}                              & UAR   & WAR   & UAR   & WAR   & UAR   & WAR   \\ \midrule
\multicolumn{1}{c|}{\multirow{3}{*}{Classifier-based}} &
  Linear Probe &
  45.46 &
  57.40 &
  32.47 &
  43.72 &
  30.74 &
  42.95 \\
\multicolumn{1}{c|}{} & Fully Fine-Tuning (w/o TM) & 55.70 & 68.41 & 39.64 & 50.77 & 37.53 & 50.48 \\
\multicolumn{1}{c|}{} & Fully Fine-Tuning (w/ TM)  & 58.28 & 70.27 & 40.55 & 51.22 & 38.39 & 50.92 \\ \midrule
\multicolumn{1}{c|}{\multirow{5}{*}{\begin{tabular}[c]{@{}c@{}}Text-based\\ (Classifier-free)\end{tabular}}} &
  Zero-shot CLIP \cite{radford2021learning} & 23.34 & 20.07 & 20.99 & 17.09 & 18.42 & 19.16 \\
\multicolumn{1}{c|}{} & Zero-shot FaRL \cite{zheng2022general} & 23.14 & 31.54 & 21.67 & 25.65 & 14.18 & 11.78 \\
\multicolumn{1}{c|}{} & CoOp \cite{zhou2022learning}   & 44.98 & 56.68 & 31.72 & 42.55 & 30.79 & 42.77 \\
\multicolumn{1}{c|}{} & Co-CoOp \cite{zhou2022conditional}  & 46.80 & 57.52 & 32.91 & 44.25 & 30.81 & 43.23 \\
\multicolumn{1}{c|}{} & DFER-CLIP (w/o TM) (Ours) & 57.39 & 69.00 & 40.64 & 50.92 & 38.51 & 50.91 \\
\multicolumn{1}{c|}{} & DFER-CLIP (w TM) (Ours)   & \textbf{59.61} & \textbf{71.25} & \textbf{41.27} & \textbf{51.65} & \textbf{39.89} & \textbf{52.55} \\ \bottomrule
\end{tabular}
\end{center}
\caption{Evaluation of different training strategies. Linear Probe: Adding a linear classifier to the CLIP image encoder and optimizing the parameters of the linear layer with the image encoder fixed. Fully fine-tuning: fine-tuning both the linear layer and the CLIP image encoder. Zero-shot CLIP/FaRL: Using the prompt ``an expression of [class]''. CoOp \& Co-CoOp: Using the prompt ``[learnable prompt] [class]''. The image encoder in FaRL is ViT-B/16-based and the rest are all ViT-B/32-based. TM stands for the temporal model.}
\label{tab2}
\end{table}
\begin{table}[!t]
\begin{center}
\small
\begin{tabular}{@{}cc|cc|cc|cc@{}}
\toprule
\multicolumn{2}{c|}{\multirow{2}{*}{Prompts}}                                                                                     & \multicolumn{2}{c|}{DFEW}       & \multicolumn{2}{c|}{FERV39k}    & \multicolumn{2}{c}{MAFW}        \\ \cmidrule(l){3-8} 
\multicolumn{2}{c|}{}                                                                                                             & UAR            & WAR            & UAR            & WAR            & UAR            & WAR            \\ \midrule
\multicolumn{1}{c|}{\multirow{4}{*}{\begin{tabular}[c]{@{}c@{}}w/o\\ TM\end{tabular}}} & a photo of {[}Class{]}                   & 56.21          & 68.44          & 39.44          & 49.94          & 37.91          & 50.87          \\
\multicolumn{1}{c|}{}                                                                  & an expression of {[}Class{]}             & 56.16          & 68.73          & 39.28          & 50.41          & 37.71          & 51.08          \\
\multicolumn{1}{c|}{}                                                                  & {[}Learnable Prompt{]} {[}Class{]}       & 57.37          & 68.86          & 40.42          & 50.50          & 38.01          & 50.81          \\
\multicolumn{1}{c|}{}                                                                  & {[}Learnable Prompt{]} {[}Descriptors{]} & 57.39          & 69.00          & 40.64          & 50.92          & 38.51          & 50.91          \\ \midrule
\multicolumn{1}{c|}{\multirow{2}{*}{\begin{tabular}[c]{@{}c@{}}w/ \\ TM\end{tabular}}} & {[}Learnable Prompt{]} {[}Class{]}       & 58.28          & 70.29          & 40.60          & 51.18          & 39.64          & 51.21          \\
\multicolumn{1}{c|}{}                                                                  & {[}Learnable Prompt{]} {[}Descriptors{]} & \textbf{59.61} & \textbf{71.25} & \textbf{41.27} & \textbf{51.65} & \textbf{39.89} & \textbf{52.55} \\ \bottomrule
\end{tabular}
\end{center}
\caption{Evaluation of different prompts. We set the number of the learnable context prompts as 8 and with or without the temporal model for comparison.}
\label{tab3}
\end{table}
\begin{table}[t]
\begin{center}
\small
\begin{tabular}{@{}c|c|cc|cc|cc@{}}
\toprule
\multirow{2}{*}{\begin{tabular}[c]{@{}c@{}}Description\\ Position\end{tabular}} &
  \multirow{2}{*}{\begin{tabular}[c]{@{}c@{}}Class\\ Specific\end{tabular}} &
  \multicolumn{2}{c|}{DFEW} &
  \multicolumn{2}{c|}{FERV39k} &
  \multicolumn{2}{c}{MAFW} \\ \cmidrule(l){3-8} 
       &              & UAR            & WAR            & UAR            & WAR            & UAR            & WAR            \\ \midrule
Middle & \XSolidBrush & 58.91          & 71.19          & 40.36          & 51.22          & 39.55          & 52.54          \\
Middle & \Checkmark   & 59.24          & \textbf{71.33} & 40.71          & 51.32          & 39.26          & 52.19          \\ \midrule
End    & \XSolidBrush & 58.97          & 71.12          & 40.50          & 51.21          & 39.46          & 52.45          \\
End    & \Checkmark   & \textbf{59.61} & 71.25          & \textbf{41.27} & \textbf{51.65} & \textbf{39.89} & \textbf{52.55} \\ \bottomrule
\end{tabular}
\end{center}
\caption{Evaluation of different settings for learnable context. All the methods adopt 8 learnable context tokens, with the temporal model.}
\label{tab4}
\end{table}

\textbf{\textit{Evaluation of Different Training Strategies}}: The final recognition result in our DFER-CLIP is obtained by computing the similarities between visual embedding and all the textual class-level embeddings, which is different from the conventional classifier-based training strategies. Therefore, we compare DFER-CLIP with Linear Probe and Fully Fine-tuning methods. The results in Tab. \ref{tab2} show that our method outperforms Fully Fine-tuning in UAR by 3.91\%, 1.63\%, and 2.36\%, and in WAR by 2.84\%, 0.88\%, and 2.07\% on DFER, FERV39k, and MAFW datasets, respectively. Even without the temporal model, our method is better than all the classifier-based methods. We also add a temporal model for the Fully Fine-tuning strategy, the results demonstrate our method is still superior to it. We also compare our method with zero-shot CLIP \cite{radford2021learning} and zero-shot FaRL \cite{zheng2022general}, in which FaRL is pre-trained on the large-scale visual-language face data. The results in Tab. \ref{tab2} show that fine-tuning the image encoder can improve the performance remarkably. CoOp \cite{zhou2022learning} and Co-CoOp \cite{zhou2022conditional} are all using the learnable context, in which the Co-CoOp also add projected image features into the context prompt. The comparison between our method without a temporal model and these two methods in Tab. \ref{tab2} show the effectiveness of our strategy.

\begin{table}[!t]
\small
\begin{center}
\begin{tabular}{@{}c|cc|cc|cc@{}}
\toprule
\multirow{2}{*}{Methods} & \multicolumn{2}{c|}{DFEW}       & \multicolumn{2}{c|}{FERV39k}    & \multicolumn{2}{c}{MAFW}        \\ \cmidrule(l){2-7} 
                                     & UAR   & WAR   & UAR   & WAR   & UAR         & WAR         \\ \midrule
C3D \cite{tran2015learning}          & 42.74 & 53.54 & 22.68 & 31.69 & 31.17       & 42.25       \\
P3D \cite{qiu2017learning}           & 43.97 & 54.47 & 23.20 & 33.39 & -           & -           \\
I3D-RGB \cite{carreira2017quo}       & 43.40 & 54.27 & 30.17 & 38.78 & -           & -           \\
3D ResNet18 \cite{hara2018can}       & 46.52 & 58.27 & 26.67 & 37.57 & -           & -           \\
R(2+1)D18 \cite{tran2018closer}      & 42.79 & 53.22 & 31.55 & 41.28 & -           & -           \\
ResNet18-LSTM \cite{he2016deep,hochreiter1997long}  & 51.32 & 63.85 & 30.92 & 42.95 & 28.08       & 39.38       \\
ResNet18-ViT \cite{he2016deep,dosovitskiy2020image} & 55.76 & 67.56 & 38.35 & 48.43 & {\ul 35.80} & 47.72       \\
EC-STFL \cite{jiang2020dfew} [MM'20]         & 45.35 & 56.51 & -     & -     & -       & -           \\
Former-DFER \cite{zhao2021former} [MM'21]    & 53.69 & 65.70 & 37.20 & 46.85 & 31.16   & 43.27       \\
NR-DFERNet \cite{li2022nr} [arXiv'22]        & 54.21 & 68.19 & 33.99 & 45.97 & -       &             \\
DPCNet \cite{wang2022dpcnet} [MM'22]         & 57.11 & 66.32 & -     & -     & -       & -           \\
T-ESFL \cite{liu2022mafw}  [MM'22]           & -     & -     & -     & -     & 33.28   & {\ul 48.18} \\
EST \cite{liu2023expression} [PR'23]         & 53.94 & 65.85 & -     & -     & -       & -           \\
LOGO-Former \cite{ma2023logo} [ICASSP'23]    & 54.21 & 66.98 & 38.22 & 48.13 & -       & -           \\
IAL \cite{li2023intensity} [AAAI'23]         & 55.71 & 69.24 & 35.82 & 48.54 & -           & -           \\
CLIPER  \cite{li2023cliper} [arXiv'23]       & {\ul 57.56}    & {\ul 70.84}    & {\ul 41.23}    & {\ul 51.34}    & - & - \\ 
M3DFEL \cite{wang2023rethinking} [CVPR'23]   & 56.10 & 69.25 & 35.94 & 47.67 & - & - \\
AEN \cite{lee2023frame} [CVPRW'23]           & 56.66 & 69.37 & 38.18 & 47.88 & - & - \\
\midrule
DFER-CLIP (Ours) & \textbf{59.61} & \textbf{71.25} & \textbf{41.27} & \textbf{51.65} & \textbf{39.89} & \textbf{52.55} \\ \bottomrule
\end{tabular}
\end{center}
\caption{Comparison with the state-of-the-art methods.}
\label{tab5}
\end{table}

\textbf{\textit{Evaluation of Different Prompts}}: In a supervised learning setting, V-LP models differ from conventional classification models in that they can be prompted to design classifier-free models for prediction. This makes prompt engineering a critical aspect of adapting V-LP models for downstream tasks. We compare our method with two kinds of manually designed prompts: ``a photo of [class]'' and  ``an expression of [class]''. Results in Tab. \ref{tab3} show that our method outperforms manually designed prompts on both DFEW and FERV39k datasets, but has a slightly lower WAR than the prompt of ``an expression of [class]'' on the MAFW dataset. We note that the MAFW dataset comprises eleven facial expression classes and that the video samples are imbalanced, with the proportions of the ``contempt'', ``helplessness'', and ``disappointment'' expressions accounting for only 2.57\%, 2.86\%\, and 1.98\%, respectively. The results in Tab. \ref{tab3} demonstrate that the learning-based context consistently achieves the best WAR results. Furthermore, our method outperforms using the prompt of the class name with the learnable context approach, which indicates the effectiveness of using descriptions.

In our DFER-CLIP, the description is at the end of the prompt and the learnable context is class-specific, namely, each description will have its own learned context. We also conduct experiments with different settings: a) put the description in the middle of the learnable context, and b) all the descriptions share the same learned context. The results on three datasets are shown in Tab. \ref{tab4}. The results indicate that positioning the description tokens to the end gains more improvement. Moreover, for the end position, adopting the class-specific is always better than a shared context prompt, achieving the best results.

\subsection{Comparison with State-of-the-Art Methods}
In this section, we compare our results with several state-of-the-art methods on the DFEW, FERV39k, and MAFW benchmarks. Consistent with previous methods \cite{jiang2020dfew,zhao2021former,ma2023logo,li2023intensity,li2023cliper,liu2022mafw}, the experiments on DFEW and MAFW are conducted under 5-fold cross-validation and use training and test set on FERV39k. Furthermore, we train models three times with different random seeds and then use the average for more stable and reliable results. The comparative performance in Tab. \ref{tab5} demonstrates that the proposed DFER-CLIP outperforms the compared methods both in UAR and WAR. Specifically, compared with the previous best results, our method shows a UAR improvement of 2.05\%, 0.04\%, and 4.09\% and a WAR improvement of 0.41\%, 0.31\%, and 4.37\% on DFEW, FERV39k, and MAFW, respectively. It should be pointed out that FERV39k is the current largest DFER benchmark with 38,935 videos. Given this substantial scale, making significant enhancements becomes a formidable task.  

\section{Conclusion}
This paper presents a novel visual-language model called DFER-CLIP for in-the-wild dynamic facial expression recognition.  In the visual part, based on the CLIP image encoder, a temporal model consisting of several Transformer Encoders was introduced for modelling the temporal facial expression features. In the textual part, the expression descriptors related to facial behaviour were adopted as the textual input to capture the relationship between facial expressions and their underlying facial behaviour, in which the expression descriptors were generated by large language models like ChatGPT. The learnable contexts for these descriptors were also designed to help the model learn relevant context information for each expression during training. Extensive experiments demonstrate the effectiveness of each component in DFER-CLIP. Moreover, the proposed method achieves state-of-the-art results on three benchmarks.

\section*{Acknowledgements}
Zengqun Zhao is funded by Queen Mary Principal's PhD Studentships, and this work is supported by the EU H2020 AI4Media No. 951911 project. We thank Niki Foteinopoulou and James Oldfield for the valuable discussion.
\bibliography{reference}

\appendix
\newpage

\begin{centering}
    \Large \textbf{Prompting Visual-Language Models for Dynamic Facial Expression Recognition (Appendix)} \\
\end{centering}

\renewcommand{\thesubsection}{\Alph{subsection}}
\subsection{Facial Expression Descriptions}

\begin{table}[h]
\begin{center}
\small
\begin{tabular}{c|c}
\toprule
\textbf{Expressions} & \textbf{Descriptors}                                                                                                         \\ \midrule
happiness            & \begin{tabular}[c]{@{}c@{}}a smiling mouth, raised cheeks, wrinkled eyes, and arched eyebrows.\end{tabular}                 \\ \midrule
sadness              & \begin{tabular}[c]{@{}c@{}}tears, a downward turned mouth, \\ drooping upper eyelids,and a wrinkled forehead.\end{tabular}     \\ \midrule
neutral              & \begin{tabular}[c]{@{}c@{}}relaxed facial muscles, a straight mouth, \\ a smooth forehead, and unremarkable eyebrows.\end{tabular}           \\ \midrule
anger                & \begin{tabular}[c]{@{}c@{}}furrowed eyebrows, narrow eyes, \\ tightened lips, and flared nostrils.\end{tabular}                \\ \midrule
surprise             & \begin{tabular}[c]{@{}c@{}}widened eyes, an open mouth, raised eyebrows, \\ and a frozen expression.\end{tabular}              \\ \midrule
disgust              & \begin{tabular}[c]{@{}c@{}}a wrinkled nose, lowered eyebrows, \\ a tightened mouth, and narrow eyes.\end{tabular}              \\ \midrule
fear                 & \begin{tabular}[c]{@{}c@{}}raised eyebrows, parted lips, \\ a furrowed brow, and a retracted chin.\end{tabular}                \\ \midrule
contempt             & \begin{tabular}[c]{@{}c@{}}one side of its mouth raised, \\ one eyebrow lower and one raised, \\ narrowed eyes, and a raised chin.\end{tabular} \\ \midrule
anxiety              & \begin{tabular}[c]{@{}c@{}}a tensed forehead, tightly pressed lips, \\ pupil dilation, and tensed facial muscles.\end{tabular} \\ \midrule
helplessness         & \begin{tabular}[c]{@{}c@{}}drooping eyebrows, a downward gaze, \\ a downturned mouth, and lacking expression.\end{tabular}     \\ \midrule
disappointment       & \begin{tabular}[c]{@{}c@{}}a downturned mouth, lowered eyebrows, \\ narrowed eyes, and a sighing face.\end{tabular}            \\ \bottomrule
\end{tabular}
\end{center}
\end{table}

\subsection{Additional Ablation Studies}
The descriptions used in our method are class-level instead of sample-level, and our point is to create discriminative class-level text embedding for supervising the visual part. However, we should admit that the unique description has limitations. To this end, we also conducted experiments with prompt ensembles; the results are shown in Tab. \ref{tabA} and indicate that employing more descriptions is beneficial.

In our method, we prompt a large language model such as ChatGPT to automatically generate useful visual descriptors for each facial expression with the process described in Section 3.3. Furthermore, we also investigated the effect of the different descriptions. Specifically, we study the effect of the different numbers of the facial-action-unit-level descriptors. We select the top-2, top-4, and top-6 descriptors generated from the LLMs -- the results are shown in Tab. \ref{tabB}. We believe fewer descriptors cause the lack of correlation among different expressions and more descriptors result in diminishing the discriminative features.

\begin{table}[h]
\renewcommand\thetable{A}
\begin{center}
\caption{Evaluation of the prompt ensembling.}
\label{tabA}
\footnotesize
\begin{tabular}{@{}c|cc|cc|cc@{}}
\toprule
\multirow{2}{*}{\begin{tabular}[c]{@{}c@{}}descriptions\\ number\end{tabular}} & \multicolumn{2}{c|}{DFEW} & \multicolumn{2}{c|}{FERV39k} & \multicolumn{2}{c}{MAFW} \\ \cmidrule{2-7} 
  & UAR            & WAR            & UAR            & WAR            & UAR            & WAR            \\ 
  \midrule
1 & 59.61          & 71.25          & \textbf{41.27} & \textbf{51.65} & 39.89          & 52.55          \\
2 & \textbf{60.42} & \textbf{72.01} & 40.84          & 51.60           & \textbf{40.42} & \textbf{52.92} \\ 
\bottomrule
\end{tabular}
\end{center}
\vspace{-1em}
\end{table}
\begin{table}[h]
\renewcommand\thetable{B}
\begin{center}
\caption{Evaluation of the different number of descriptors.}
\label{tabB}
\footnotesize
\begin{tabular}{c|cc|cc|cc}
\toprule
\multirow{2}{*}{\begin{tabular}[c]{@{}c@{}}descriptiors\\ number\end{tabular}} & \multicolumn{2}{c|}{DFEW} & \multicolumn{2}{c|}{FERV39k} & \multicolumn{2}{c}{MAFW} \\ \cmidrule{2-7} 
  & UAR            & WAR            & UAR            & WAR            & UAR            & WAR            \\ \midrule
2 & \textbf{59.65} & \textbf{71.91} & 40.79          & 51.57          & 39.65          & 52.26          \\
4 & 59.61          & 71.25          & \textbf{41.27} & \textbf{51.65} & \textbf{39.89} & \textbf{52.55} \\
6 & 59.59          & 71.87          & 40.54          & 51.46          & 39.25          & 52.37          \\ \bottomrule
\end{tabular}
\end{center}
\vspace{-1em}
\end{table}

\subsection{Confusion Matrix on Three Benchmarks}
\begin{figure}[h]
\renewcommand\thefigure{A}
    \centering
    {\label{fig_a1}\includegraphics[scale=0.13]{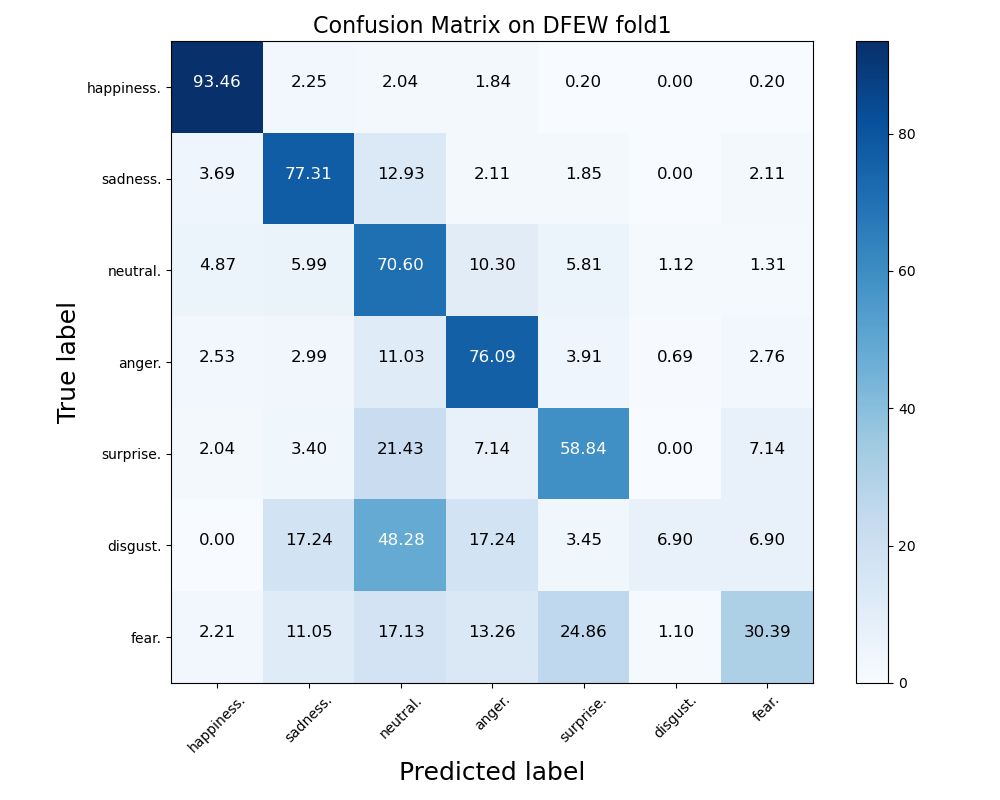}} \hspace{-6mm}
    {\label{fig_a2}\includegraphics[scale=0.13]{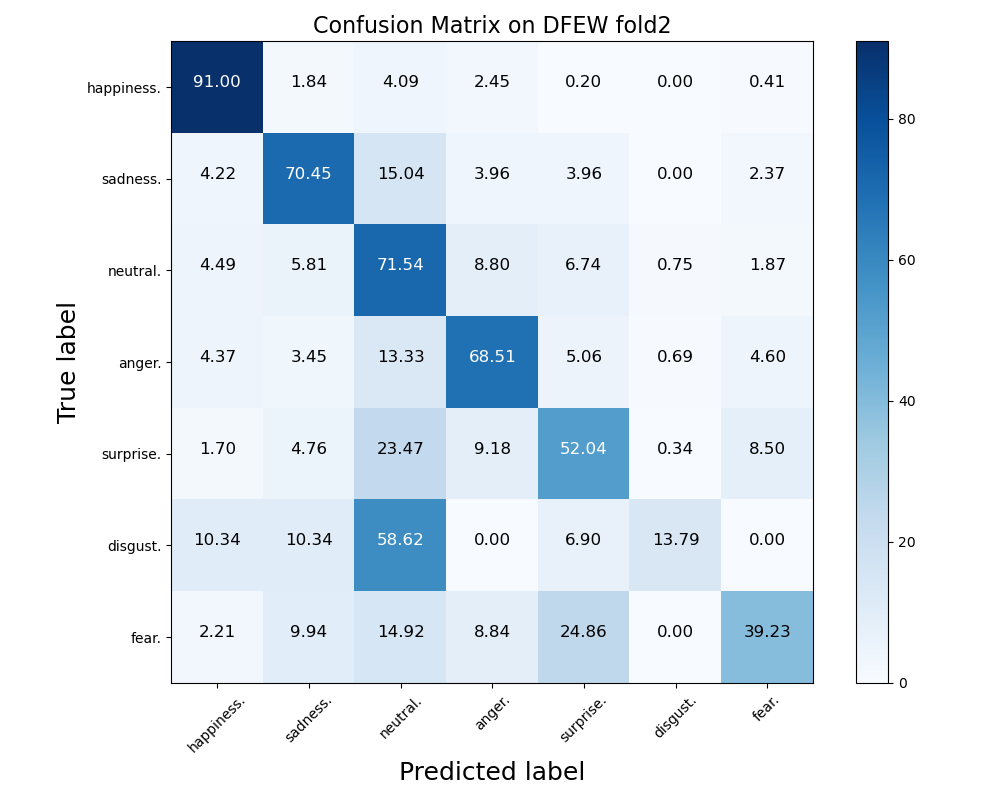}} \hspace{-6mm}
    {\label{fig_a3}\includegraphics[scale=0.13]{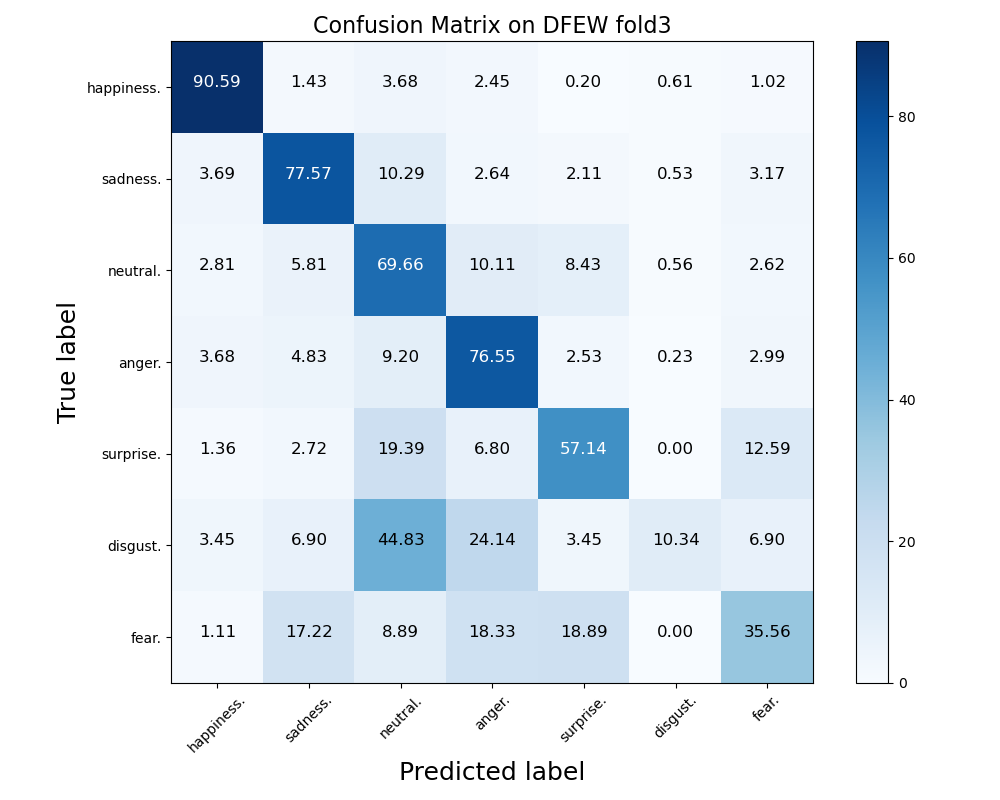}} \hspace{-6mm}
    {\label{fig_a4}\includegraphics[scale=0.13]{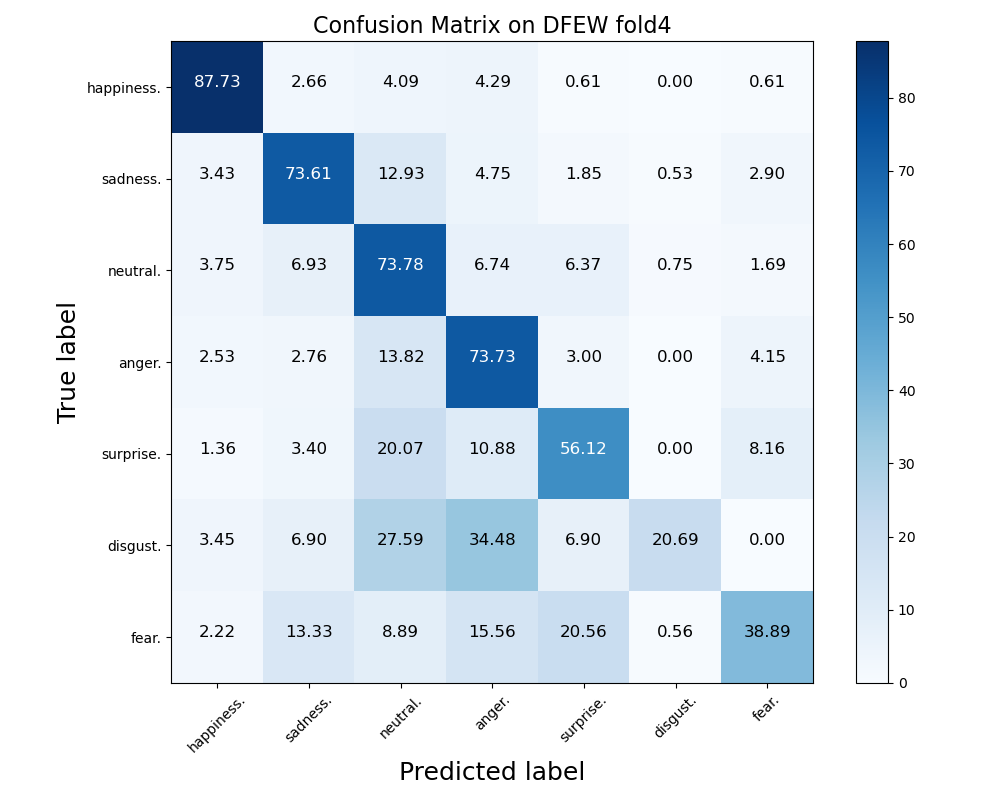}} \hspace{-6mm}
    {\label{fig_a5}\includegraphics[scale=0.13]{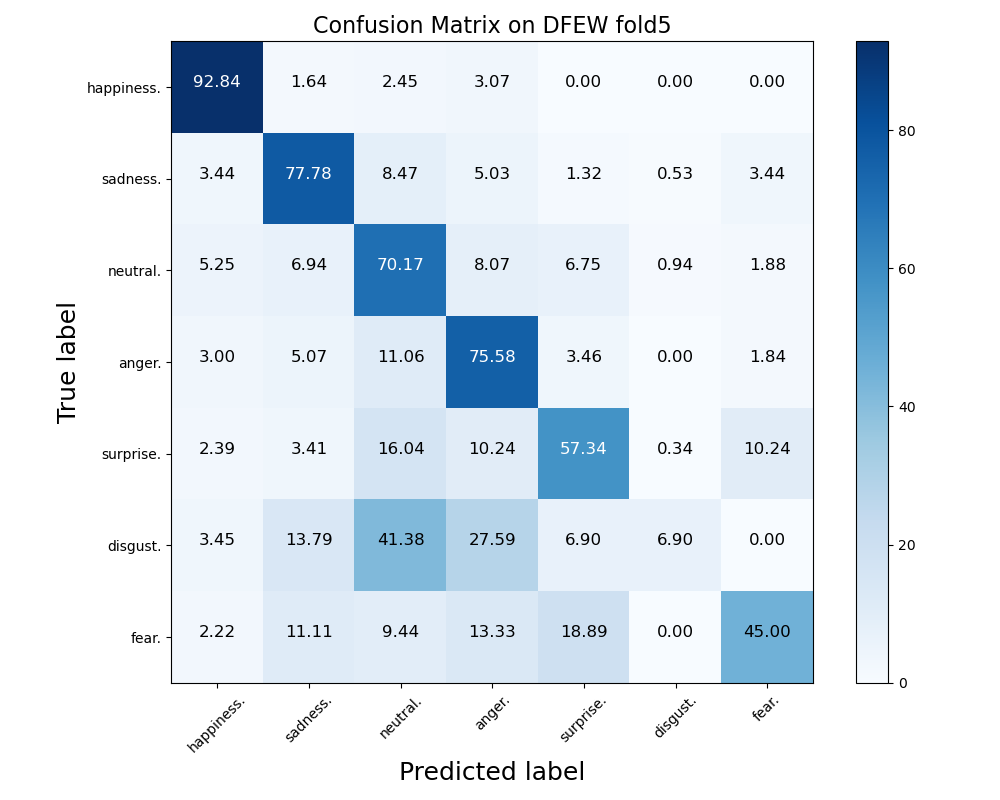}} \hspace{-6mm}
    {\label{fig_a6}\includegraphics[scale=0.13]{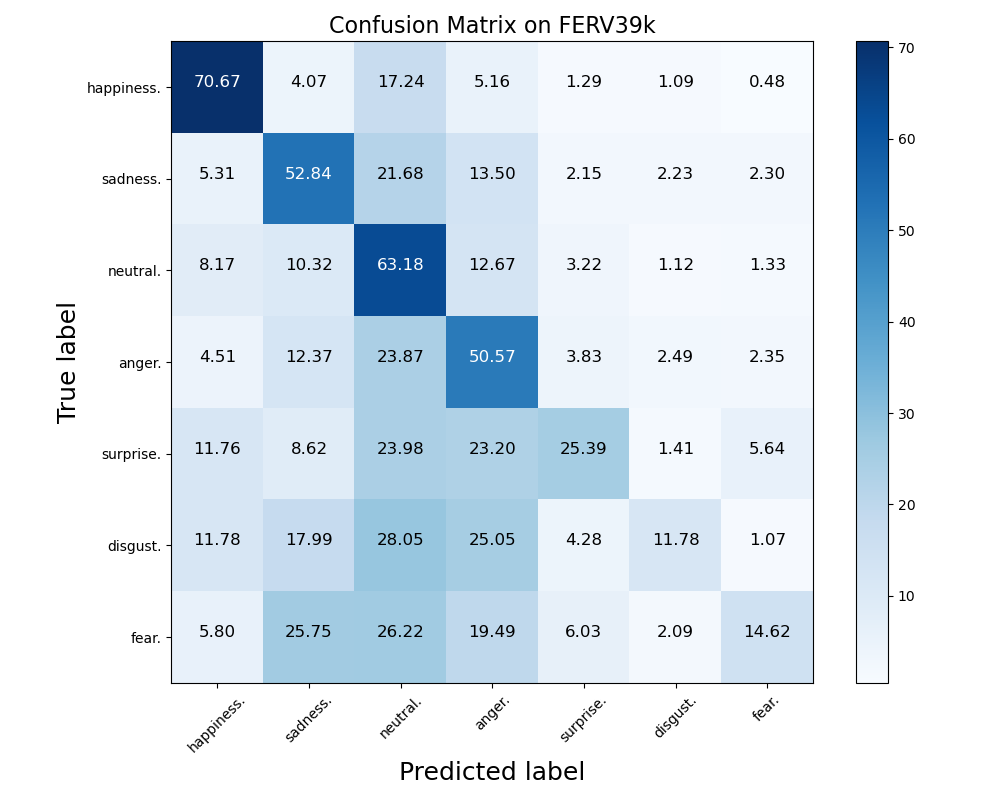}}
    {\label{fig_a7}\includegraphics[scale=0.13]{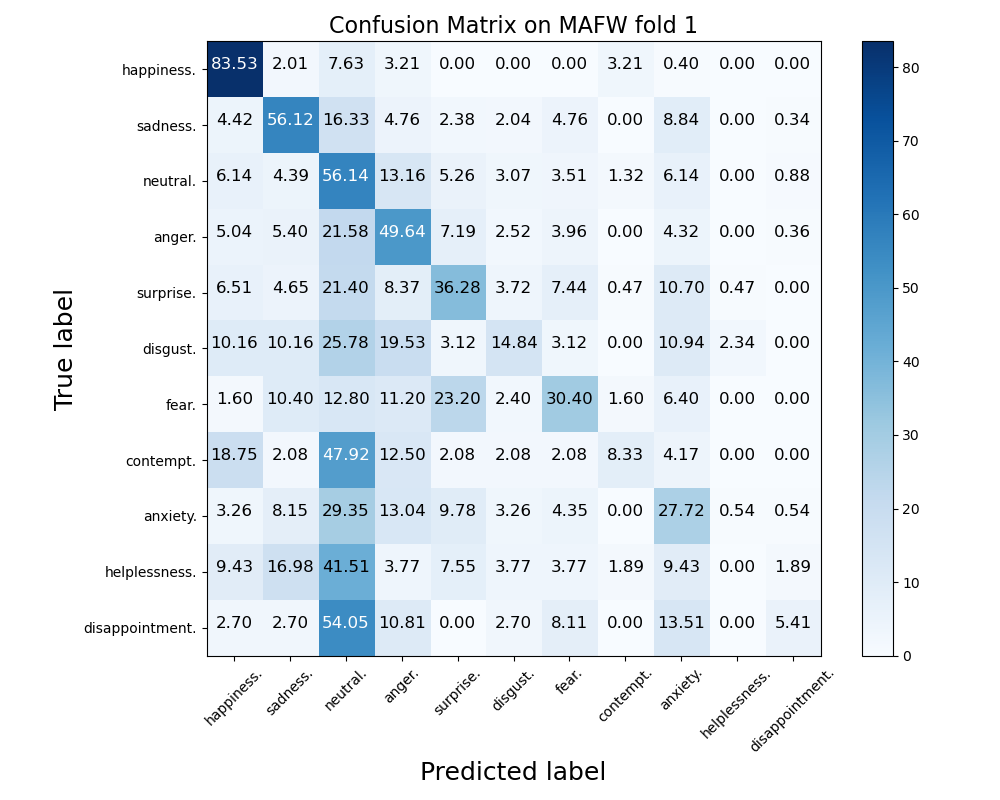}} \hspace{-6mm}
    {\label{fig_a8}\includegraphics[scale=0.13]{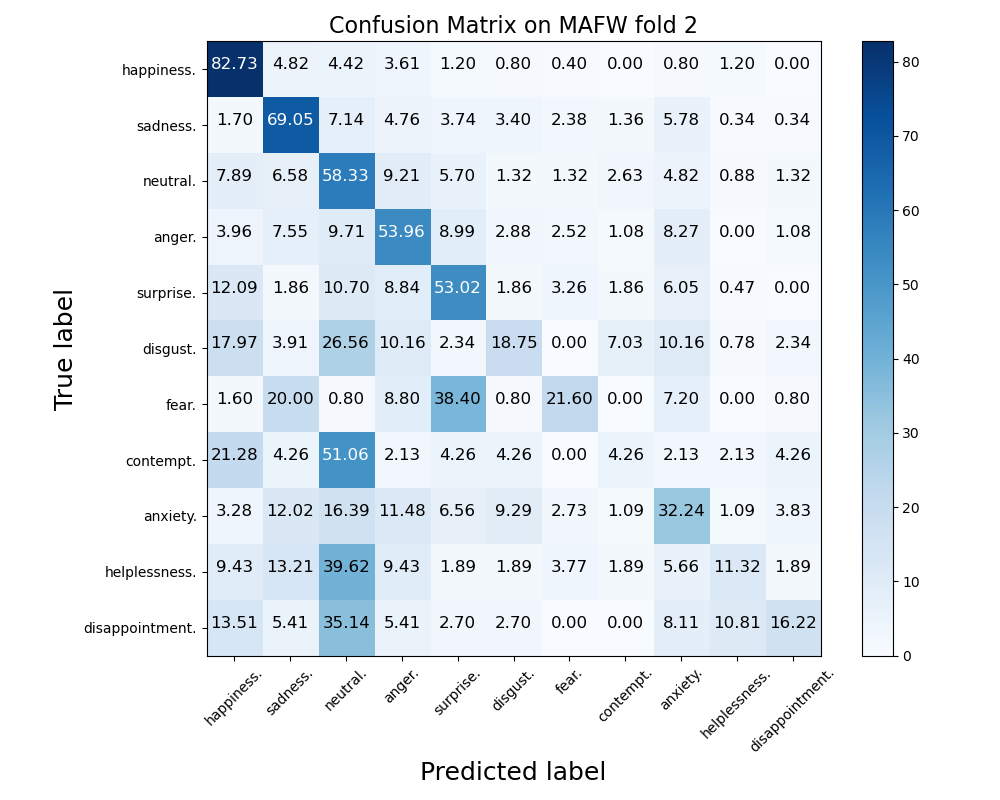}} \hspace{-6mm}
    {\label{fig_a9}\includegraphics[scale=0.13]{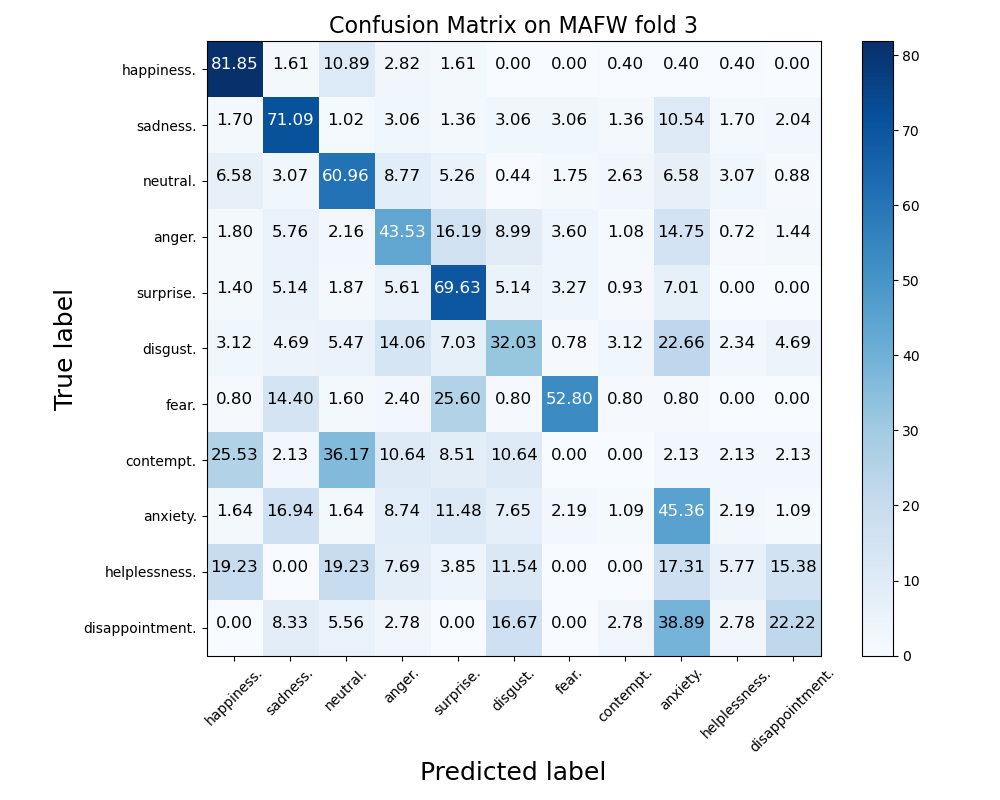}} \hspace{-6mm}
    {\label{fig_a10}\includegraphics[scale=0.13]{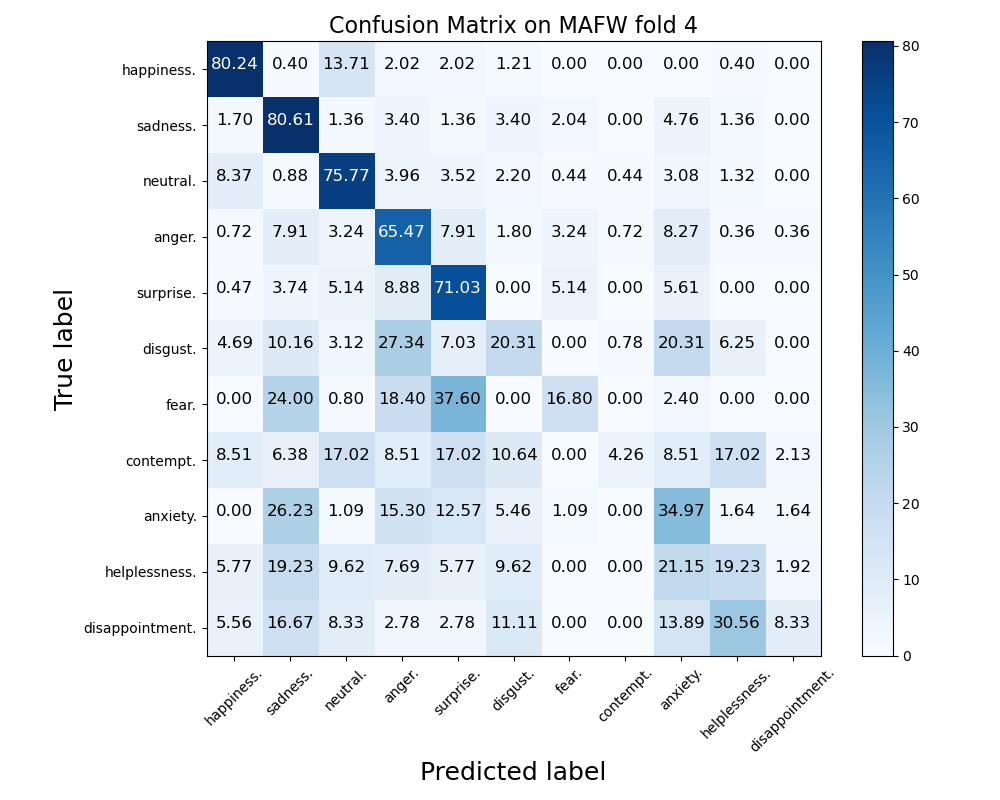}} \hspace{-6mm}
    {\label{fig_a11}\includegraphics[scale=0.13]{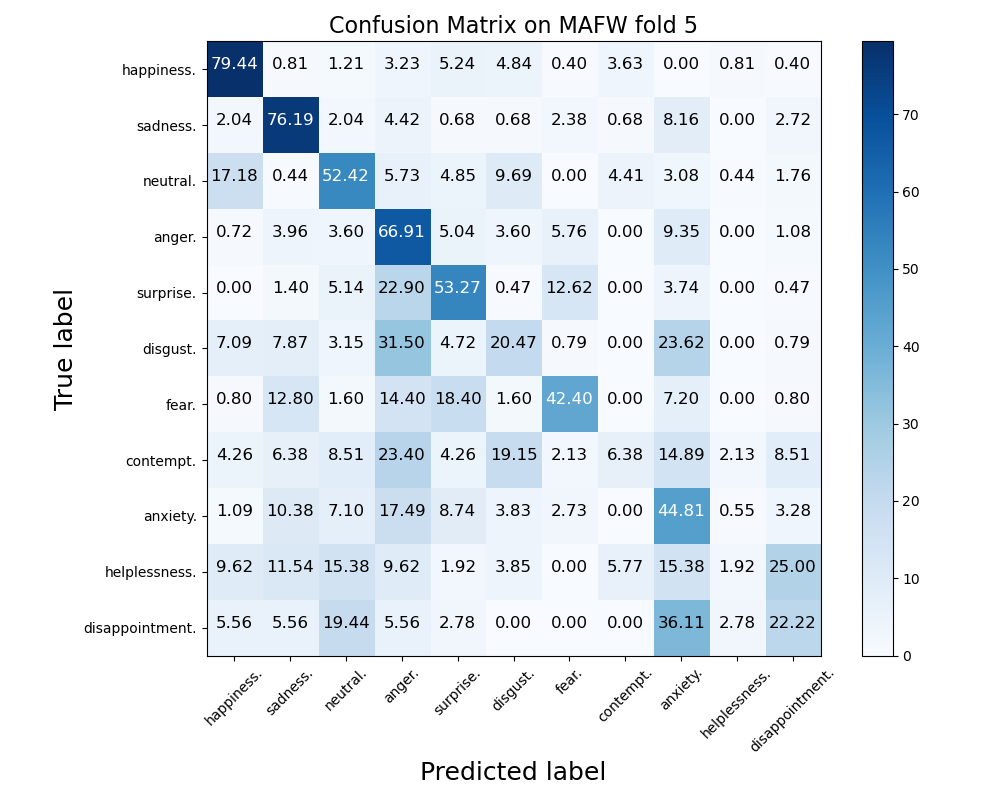}} \hspace{-6mm}
    \caption{Confusion matrix on 5-fold DFEW, FERV39k test set and 5-fold MAFW.}
\end{figure}

\end{document}